\pdfoutput=1
\documentclass[11pt]{article}

\usepackage{acl}
\usepackage{hyperref}
\usepackage{times}
\usepackage{latexsym}
\usepackage{multirow}
\usepackage{booktabs}
\usepackage{microtype}
\usepackage{multirow}
\usepackage{latexsym}
\usepackage{booktabs}
\usepackage{courier}
\usepackage{amsmath}
\usepackage{booktabs} % For better table formatting
\usepackage{siunitx} % For aligning numbers by decimal point
\usepackage{subfig}
\usepackage{placeins}
\usepackage{tikz}
\usepackage{float}
\usepackage{algorithm2e}
\RestyleAlgo{ruled}

\usepackage[T1]{fontenc}
\usepackage[utf8]{inputenc}

\usepackage{microtype}

\usepackage{inconsolata}

\title{CRCL at SemEval-2024 Task 2: Simple prompt optimizations}

\author{Clément Brutti-Mairesse \\
  CRCL / Lyon, France \\
  \texttt{\small clement.bruttimairesse@lyon.unicancer.fr} \\\And
  \textbf{Loïc Verlingue} \\
  CLB/CRCL / Lyon, France \\
  \texttt{\small loic.verlingue@lyon.unicancer.fr} \\}

\begin{document}
\maketitle
\begin{abstract}

We present a baseline for the SemEval 2024 task 2 challenge, whose objective is to ascertain the inference relationship between pairs of clinical trial report sections and statements.

We apply prompt optimization techniques with LLM Instruct models provided as a Language Model-as-a-Service (LMaaS).

We observed, in line with recent findings, that synthetic CoT prompts significantly enhance manually crafted ones.

The source code is available at this GitHub repository \href{https://github.com/ClementBM-CLB/semeval-2024}{github.com/ClementBM-CLB/semeval-2024}.

\end{abstract}

\section{Introduction}

Since the introduction of large pre-trained transformer models such as GPT-3.5, released in early 2022, foundational models have begun to be utilized widely. While BERT-like models have proven to be effective in various NLP tasks such as Named Entity Recognition \cite{devlin_bert_2019}, scaling up the number of parameters in transformer models not only enhances their capabilities but also endows them with new abilities not seen in smaller models \cite{zhao_survey_2023}. These capabilities are particularly evident in natural language inference tasks, where the model must deduce the veracity of two given texts \cite{zhong_can_2023}.

LLMs, gaining popularity for their reasoning capabilities, still face trustworthiness concerns, crucial in the medical domain where decisions affect lives. Medical devices must exhibit reliability and undergo rigorous testing before they are brought to market. SemEval 2024 \cite{jullien-etal-2023-semeval, jullien-etal-2024-semeval} focuses on assessing NLI system robustness, coherence, and accuracy, particularly LLMs prone to shortcut learning, factual discrepancies, and performance degradation from word distribution shifts \cite{liu_evaluating_2023}.

Fine-tuning, while effective for task and domain adaptation, demands excessive resources in the case of large language models (LLMs). In the medical field, data is highly sensitive and protected by privacy regulations. Therefore, applying fine-tuning techniques to such sensitive data would imply that medical centers have readily available on-premise infrastructure \cite{sun_make_2023}. Considering these limitations, we investigate hard prompt optimization techniques such as Chain-of-Thought prompting \cite{wei_chain--thought_2023}. Acknowledging the in-context learning (ICL) as an indirect method of fine-tuning, we also explored in-context learning strategies \cite{dai_why_2023}. Among them, we were particularly inspired by MedPrompt, a promising composite prompting method applied to medical datasets, which achieved a 27\% reduction in error rates on MedQA \cite{nori_can_2023}.

Following the SemEval 2024 task 7 \cite{jullien-etal-2023-nli4ct}, SemEval 2024 task 2 focuses on identifying the inference relationship (entailment vs. contradiction) between Clinical Trial Report (CTR) statement pairs. These statements and the supporting evidence are crafted by individuals with expertise in the clinical domain, including clinical trial organizers and research oncologists. The clinical trials information is sourced from the \footnote{https://clinicaltrials.gov/}{clinicaltrials.gov} website (maintained by the NIH). We have evaluated three LLM prompting methods to address this task.  

\section{Methods}

\subsection{Tasks}
The challenge involves analyzing a statement alongside one or two clinical trial reports to ascertain if the statement logically follows from the information presented in the clinical trial. Typically, a statement is a concise text averaging 19.5 words and may contain one or several claims pertaining to the clinical trial. It refers to one of four sections of the clinical trial report: Adverse Events, Eligibility Criteria, Results, or Interventions. Each section represents a distinct part of the clinical trial documentation as recorded in the clinicaltrials.gov database. The text from these sections has an average length of 265 words.

\begin{figure}
    \centering
    \includegraphics[scale=0.74]{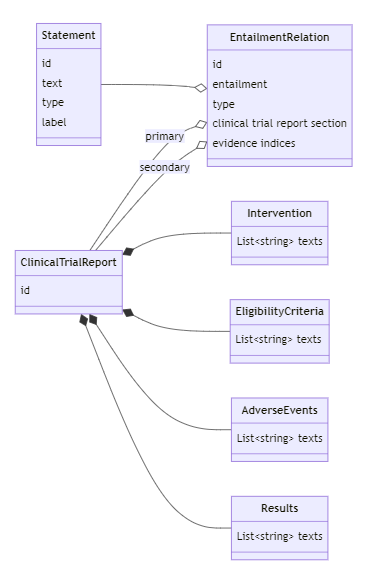}
    \caption{SemEval 2024 dataset data model}
\end{figure}

For the purpose of evaluating this task, the evaluation dataset was generated to allow us to assess the reliability (faithfulness) and consistency of the inference predictions. This was achieved by paraphrasing the text to retain the same meaning, as well as by making minor alterations to the text that change the inference relationship.

\subsection{Prompting}

We explored three prompting optimization techniques: 1) OPRO approach, which iterates over labeled examples to determine the most effective instruction \cite{yang_large_2023}, 2) self-generated chain of thought \cite{kojima_large_2023}, 3) in-context learning (ICL) strategy by incorporating one example for one-shot prompting \cite{nori_can_2023}. 

\subsection{OPRO optimization}

The OPRO technique exploits the model's capability to generate prompts based on a few exemplars and previous instructions.

In essence, the model is tasked with creating prompt instructions that are intended to solve the given problems. While this method enables the discovery of the most suitable instructions for each set, it still demands extensive resources due to its iterative optimization process. For this reason, we apply this technique to only a subset of representative examples from the development dataset.

\begin{algorithm}
\caption{OPRO prompt optimization}
\KwData{$N$ samples, $M$ test samples and $P$ instructions and their F1 scores}
\KwResult{$P$ instructions}

\For{10 times}{
    Format the $P$ instructions and $N$ samples as a context $C$ for the LLM\\
    Generate instruction with the LLM and context $C$\\
    \For{$M$ test samples}{
        Format the instruction and the test sample as a context\\
        Generate prediction with the LLM\\
    }
    Calculate the F1 score for the generated instruction\\
    Add the new instruction to the $P$ list if its F1 score is greater than the lower instruction's score of the list\\
}

\label{algo:OPRO-prompt}
\end{algorithm}

\subsection{Zero-shot Chain-of-Thought prompt}

Unlike the previous method, which constrained instructions based on the type and section of the sample, we allowed the model to generate a chain of thought reasoning using a task-agnostic meta-prompt.

\begin{algorithm}
\caption{Zero-shot Chain-of-Thought prompt}
\KwData{$N$ samples}
\KwResult{$N$ predictions}

\For{$N$ samples}{
    Format the $N$ samples as a context $C_{reasoning}$\\
    Generate chain-of-thought with the LLM and the context $C_{reasoning}$\\
    Format the generated chain-of-thought with the sample and the formating instruction\\
    Generate the prediction with the LLM and the context $C_{formating}$\\
}
\label{algo:zeroshot-cot-prompt}
\end{algorithm}

The model first generate a CoT reasoning to answer the question. Then, given the previous, it is prompted to generate a conclusion and provide the final answer—whether it entails or contradicts—in a standardized json format (algorithm \ref{algo:zeroshot-cot-prompt}).
See the figure \ref{figure:cot-prompting-sample} in appendix for a detailed example.

\subsection{Dynamic one-shot Chain-of-Thought prompt}

We hypothesized that selecting one meaningful example from a set (statement, clinical trial report) with a correct reasoning path could enhance the performance of the NLI system.

This experiment is divided into two tasks. First, we build a database of exemplars from the train dataset. Each sample corresponds to a statement and a clinical trial report section, along with its associated reasoning path (generated by the model) and predicted label. We filter the records where the model provides correct answers and index the embeddings of the statements into a vector database.

Next, for each test sample, we select a sample from the train dataset that is semantically close according to the squared L2 distance defined as $d = \sum\left(s_i-s^{train}_i\right)^2$. We choose the $s^{train}$ sample with the lowest distance to the $s$ sample that has either the same type, the same section, or both, preferably.

\begin{algorithm}
\caption{Vector database building}
\KwData{$N$ training samples}
\KwResult{Vector database of statement and reasoning paths}

\For{$N$ samples}{
    Calculate the embeddings of the statement\\
    Generate prediction following the same procedure as in Algorithm \ref{algo:OPRO-prompt}\\
    If the prediction is accurate, add the embedding vector to the database\\
}

\end{algorithm}

\begin{figure}
    \centering
    \includegraphics[scale=0.46]{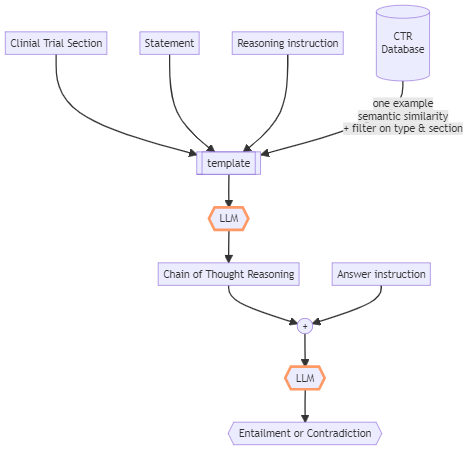}
    \caption{Dynamic one-shot prompting workflow}
    \label{figure:dynamic-oneshot}
\end{figure}

\section{Language models}

We evaluated Mixtral-8x7B-Instruct \cite{jiang_mixtral_2024}, GPT3.5 \cite{ouyang_training_2022}, Qwen-72b-chat \cite{bai_qwen_2023}, and Mistral-7B-Instruct. For all inference tasks, except instruction generation, we did not use sampling techniques.

To calculate vector embeddings, we utilized the msmarco-bert-base-dot-v5 model, in conjunction with \footnote{https://www.trychroma.com/}{chromadb} to store the embeddings in a vector database, thereby facilitating similarity score calculations using L2 norm.

\section{Evaluation metrics}

\textbf{Faithfulness} measures the extent to which a given system arrives at the correct prediction for the correct reason. This is estimated by measuring the ability of a model to correctly change its predictions when exposed to a \textbf{semantic altering} intervention.
Given $N$ statements $x_i$ in the contrast set ($C$), their respective original statements $y_i$, and model predictions $f()$ we compute faithfulness using Equation 1.
\begin{equation}
\footnotesize
    \begin{aligned}
   Faithfulness = \frac{1}{N}\sum_{1}^{N}\left| f(y_i)-f(x_i) \right|\hspace{1cm}\\ x_i\in C:\text{Label}(x_i) \neq \text{Label}(y_i), \text{ and } f(y_i) = \text{Label}(y_i) 
    \end{aligned}
\end{equation}

\textbf{Consistency} aims to measure the extent to which a given system produces the same outputs for semantically equivalent problems. Therefore, consistency is measured as the ability of a system to predict the same label for original statements and contrast statements for \textbf{semantic preserving} interventions.
Given $N$ statements $x_i$ in the contrast set ($C$), their respective original statements $y_i$, and model predictions $f()$ we compute faithfulness using Equation 2.

\begin{equation}
\footnotesize
    \begin{aligned}
   Consistency = \frac{1}{N}\sum_{1}^{N} 1 - \left| f(y_i)-f(x_i) \right| \\ x_i\in C:\text{Label}(x_i) = \text{Label}(y_i)
    \end{aligned}
\end{equation}
\newline

\section{Results}
\subsection{Main results}
Our team ranked sixth in faithfulness score, and we fell outside the top 10 for the baseline F1 score (0.70) and consistency (0.70). We observed that handcrafted prompts were generally less effective than optimized prompts or meta-prompts.

Prompting strategies were first tested on the dev dataset and then run on the test dataset. The results are shown in the table \ref{table:metrics_optimization}. Mixtral-8x7B-Instruct demonstrated the best quality-to-time ratio. The dynamic one-shot prompting achieved the highest Faithfulness score and Consistency score. While the best F1 score goes for the zero-shot CoT prompt approach. These results must be interpreted with caution because the model does not always return a well-formatted answer in JSON format.  In cases where the answered entailment label is unknown, our approach was to prioritize the contradiction label.

Because of time limitations, we had to train and assess the prompt strategy using the development dataset, which consisted of 200 samples. We solely used the training dataset to gather examples for inputting into the vector database for the one-shot prompt strategy. The execution of the entailment task on the test dataset required 20 hours for each prompting strategy. The team's outcomes for the task are presented in table \ref{table:metrics_ranking}.

\begin{table}[t]
\centering
\resizebox{\columnwidth}{!}{
\begin{tabular}{@{}lcccc@{}}
\toprule
\textbf{Model} & \textbf{Optimization} & \textbf{Base F1} & \textbf{Consistency} & \textbf{Faithfulness} \\
\midrule
Mixtral-8x7B & Zero-shot CoT & 0.70 & 0.70 & 0.87\\
Mixtral-8x7B & Dynamic one-shot & 0.60 & 0.71 & 0.89 \\
Mixtral-8x7B & OPRO & 0.59 & 0.65 & 0.81 \\
% Add more models as needed
\bottomrule
\end{tabular}%
}
\caption{Prompt optimization strategies with Mixtral-8x7B-Instruct-v0.1 on the test dataset}
\label{table:metrics_optimization}
\end{table}

\begin{table}[t]
\centering
\resizebox{\columnwidth}{!}{
\begin{tabular}{@{}lcccc@{}}
\toprule
\textbf{Ranking} & \textbf{Base F1} & \textbf{Base F1} & \textbf{Faithfulness} & \textbf{Consistency} \\
\midrule
1 & dodoodo & 0.78 (3) & 0.92 (3) & 0.81 (1)\\
2 & aryopg  & 0.78 (5) & 0.95 (2) & 0.78 (2)\\
3 & jvl & 0.78 (4) & 0.80 (13) & 0.77 (3)\\
$\cdot$ & $\cdot$ & $\cdot$ & $\cdot$ & $\cdot$ \\
17 & ClementBM  & 0.70 (18) & 0.87 (6) & 0.70 (17)\\
$\cdot$ & $\cdot$ & $\cdot$ & $\cdot$ & $\cdot$ \\
\bottomrule
\end{tabular}
}
\caption{Team ranking on the test dataset}
\label{table:metrics_ranking}
\end{table}

\subsection{Other evaluations}
We also investigated reformulation methods, such as rephrasing negative statements, paraphrasing statements to maintain the original meaning, and rewording sections of the clinical trial report \cite{cheng_adapting_2023}, we did not observe an improvement in inference accuracy (data not shown).

We observed that applying dynamic one-shot technique (F1=0.60) obtained a 10-point drop compared to the Zero-Shot CoT (F1=0.70). We also observed that implementing preprocessing steps could improve the performance of the entailment task (such as enriching the clinical trial section with additional information, transforming negative statements into positive ones, etc.). 

While experimenting with various prompt instructions to reformulate or paraphrase the statement before logical prediction on inference, we found that it didn't significantly improve performance. One detail worth mentioning is perhaps a processing step on the clinical trial report section. We observed that the model sometimes struggles to identify which paragraph of the report section matches which cohort. To address this, we explicitly added the cohort number to the subtitle of the section. All other lines of the section were concatenated without change, each separated by a newline.

\section{Conclusion}

By employing prompt optimization techniques with LLM Instruct models, we see the significant enhancement Zero-shot CoT prompts provide compared to manually crafted ones. This highlights the critical role of utilizing advanced techniques in LLM prompting to enhance inference tasks, particularly in domains like clinical trials.

\section{Acknowledgments}

This work was supported by TM2 interreg Grant from the European Regional Development Fund, TRIAL MATCH 2 (N°SYNERGIE 20023).

\newpage
\onecolumn
\bibliography{custom}

\appendix

\section{Prompt instructions}
\label{sec:appendix}

\subsection{Zero-shot CoT prompt instruction}
The following diagram illustrates with a sample from the dev dataset, how prompts are crafted. The Zero-shot CoT approach involves prompting the LLM twice.

\begin{figure}[H]
    \includegraphics[scale=0.5]{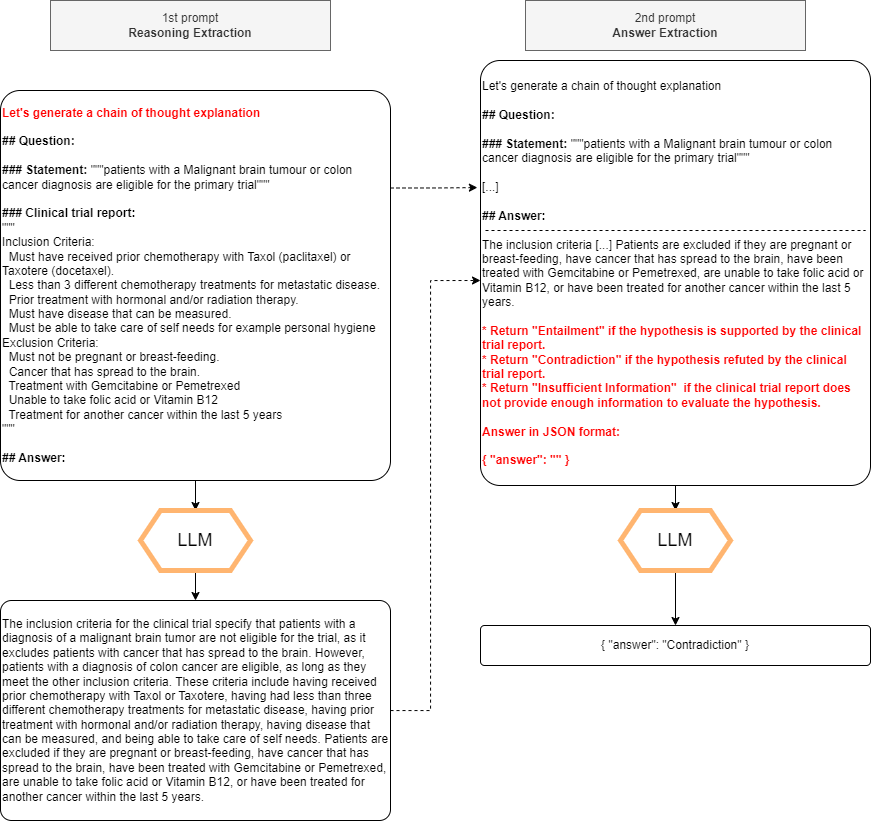}
    \caption{Zero-shot CoT prompting sample pipeline}
    \label{figure:cot-prompting-sample}
\end{figure}

\end{document}